# CENTROID-UNET: DETECTING CENTROIDS IN AERIAL IMAGES

N. Lakmal Deshapriya[1], Dan Tran[1], Sriram Reddy[1], and Kavinda Gunasekara[1]

[1]Geoinformatics Center, Asian Institute of Technology, P.O. Box 4, Klong Luang, Pathumthani 12120, Thailand,
Email: lakmal@ait.ac.th

**ABSTRACT:** In many applications of aerial/satellite image analysis (remote sensing), the generation of exact shapes of objects is a cumbersome task. In most remote sensing applications such as counting objects requires only location estimation of objects. Hence, locating object centroids in aerial/satellite images is an easy solution for tasks where the object's exact shape is not necessary. Thus, this study focuses on assessing the feasibility of using deep neural networks for locating object centroids in satellite images. Name of our model is Centroid-UNet. The Centroid-UNet model is based on classic U-Net semantic segmentation architecture. We modified and adapted the U-Net semantic segmentation architecture into a centroid detection model preserving the simplicity of the original model. Furthermore, we have tested and evaluated our model with two case studies involving aerial/satellite images. Those two case studies are building centroid detection case study and coconut tree centroid detection case study. Our evaluation results have reached comparably good accuracy compared to other methods, and also offer simplicity. The code and models developed under this study are also available in the Centroid-UNet GitHub repository: https://github.com/gicait/centroid-unet.

## 1. INTRODUCTION

Locating objects in aerial/satellite images is an important task in many remote sensing applications. In the computer vision context, objects are located either as bounding boxes enclosing objects or by estimating the exact shape of an object. In many remote sensing applications, extract shapes or the bounding boxes of objects are not essential. Some examples include, mapping building density, estimating socio-economic characteristics via building density, counting tree crowns, counting agriculture plots, and counting wild-animals etc. In this paper we are trying to locate objects just with 2D ("x" and "y") coordinates corresponding to the centroids of objects in aerial/satellite images.

On the other hand, generating ground truth data for exact shapes of objects or bounding boxes is a tedious and time consuming task (Ribera et al., 2019). Hence, by focusing more on centroid estimation, we can minimize the time that we spend for ground truth data generation. This is particularly important in large scale mapping exercises. Furthermore, after estimating centroid of objects in aerial/satellite images, feature vectors associated with the centroids can also be used to regress secondary information such as texture, shape characteristics, etc. about the object, via transfer learning if necessary.

The recent upswing in the earth observation industry has generated a large number of satellite images of the earth's surface in fine detail with frequent observations. Mapping and extracting information from them with traditional mapping methods is a challenging and time consuming task. However, recent advances in deep neural networks based methods (deep learning) provided a unique opportunity in this regard. One of deep learning architecture, currently being popular among remote sensing practitioners is the U-Net semantic segmentation architecture (Ronneberger et al., 2015). In this paper, we assess the feasibility of using classic U-Net semantic segmentation architecture (Ronneberger et al., 2015), for the centroid estimation from aerial/satellite images. We evaluate our method with two case studies which include building centroid and coconut tree centroid mapping from aerial/satellite images. Our proposed method is limited to centroid estimation of a single class. Our method can be extended to multi-class estimation with minor modification as well.

## 2. RELATED WORKS

We are not the first to use deep learning for centroid estimation. One of the well-known early publications in this line of work is known as CenterNet (Zhou et al., 2019b). This paper describes a simple single-stage, anchor free object localization method via estimating object centers. In this regard object centers are represented as small Gaussian kernels around object centroids. Sizes (standard deviation) of these Gaussian kernels were determined by the object size. In our study, we are purely focusing on the centroid without object size estimation; we can't adapt these adaptive Gaussian kernels based methods. So in our proposed method, we used a fixed size Gaussian kernel which is different from the CenterNet.

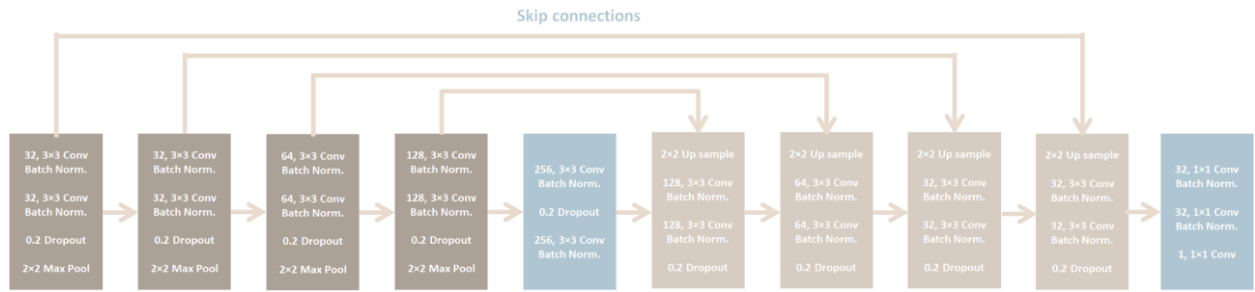

Figure 1: U-Net architecture that we have adapted.

In computer vision literature, most center / centroid estimations are involved with people, vehicle counting applications. Most of these detection methods estimate locations via bounding boxes (Dijkstra et al., 2019a). Direct estimation of the center by deep neural networks is a challenging task because the loss gradient around the center / centroid is almost zero. Moreover, networks need to estimate the center exactly to minimize the error, even close estimations do not contribute to smooth gradient. Due to this nature of the problem, direct estimation is a challenging task. To get around this issue, many approaches introduce density estimation via Gaussian kernels as in our approach (Zhang et al., 2016).This method is specific for the counting task.

Another modern innovative approach was introduced in the paper "Locating Objects Without Bounding Boxes" (Ribera et al., 2019). In this paper, they have introduced a new differentiable loss function called "The Average Hausdorff Distance". This provides a smooth gradient around the center / centroid, solving the problem of flat gradient around the center / centroid of objects. Complicated nature of this loss formulation is a disadvantage in this method. In contrast, in our study, we are assessing the feasibility of using U-Net architecture with simple mean squared error loss.

## 3. METHODOLOGY

### 3.1 Datasets

We have used 2 datasets for our experiments to evaluate the feasibility of the proposed approach, namely "Massachusetts buildings dataset" and the "Coconut trees dataset from the kingdom of Tonga (the south pacific)".

"**Massachusetts buildings dataset**" consists of 151 aerial images of 1500 x 1500 pixels, at a spatial resolution of around 1 $m^2$ per pixel (Mnih, 2013). It offers pixel wise ground truth data for two classes (buildings and non-buildings). We have resized all images in the dataset to 1280 x 1280 pixels in order to be compatible with our network architecture. And also, we have extracted centroids of ground-truth building for our experiment.

"**Coconut trees dataset from the kingdom of Tonga (the south pacific)**" were produced by WeRobotics and OpenAerialMap, the World Bank's UAVs for Disaster Resilience Program. And the dataset was captured October 2017 (World Bank, 2018). Original dataset covers around 80km² with 10 cm spatial resolution per pixel. Ground truth data include Coconut trees, Banana trees, Papaya trees and Mango trees as point data. We have extracted only coconut trees from this dataset for our experiment. And 247 images tiles of size 1280×1280 were also generated from the initial single RGB mosaic image in order to feed into the neural network.

### 3.2 Approach

U-Net semantic segmentation architecture (Ronneberger et al., 2015) was used as the backbone of our approach. The architecture of the U-Net that we have adapted for this study is shown in Figure 1. Even though we have used U-Net architecture in our approach, any semantic segmentation neural network backbone can also be used instead of the U-Net architecture.

We have modified the U-Net architecture (Figure 1) by scaling down the original architecture in order to fit into our GPU memory. We have introduced "Batch normalization" and "Dropout" regularization techniques to stabilize the network. All activation of the network was Rectified Linear Unit (ReLU), except the last layer, whose output needed to be scaled to the range of 0 to 1. So the "Sigmoid" activation function was used in the last layer.

The input images to the network were RGB aerial/satellite image tiles. First, output images are prepared as binary images (if a pixel is a centroid of an object, its value is 1; otherwise, its value is 0). Then Gaussian kernels were placed around each centroid pixel. The size (standard deviation) of the Gaussian kernel was determined empirically, such that the overlaps of Gaussian kernels are minimized. Unlike CenterNet (Zhou et al., 2019a), size (standard deviation) of the Gaussian kernel was a fixed value for each dataset (case study). In our experiments, 10 pixels and 50 pixels were chosen as the size (standard deviation) of the Gaussian kernel for building centroid detection case study and coconut tree centroid detection case study respectively. Furthermore, values of these Gaussian kernels were scaled to the range between 0 to 1, in order to be compatible with the "Sigmoid" activation function at the final layer of the U-Net. If two or more Gaussian kernels are overlapping, element-wise maximums were selected as the final value. Equations for Gaussian kernels generation and the value scaling to the range between, 0 to 1 are shown in Equation 1 and 2 respectively.

$$G = e^{-\frac{d^2}{2 \times \sigma^2}} \tag{1}$$

$$G_{scaled} = \frac{G - G_{min}}{G_{max} - G_{min}} \tag{2}$$

Where:
- $G$: Gaussian kernel
- $d$: distance from the center in pixels
- $\sigma$: size (standard deviation) of these Gaussian kernels
- $G_{scaled}$: scaled Gaussian kernel to values between 0 to 1
- $G_{min}$: minimum value of the Gaussian kernel
- $G_{max}$: maximum value of the Gaussian kernel

Input RGB satellite image tiles and output images with reconstructed Gaussian kernels around centroids were fed into the network in the training stage. In both case studies, the mean squared error (MSE) between model output and the ground truth with reconstructed Gaussian kernels around centroids was used as the loss function. Both networks are trained with the Adam optimizer (Kingma and Ba, 2015).

### 3.3 Training Details

Both networks for the two datasets were trained on a single GPU (GEFORCE GTX 1080 Ti). We have used the Keras deep learning framework with the TensorFlow backend for the implementation (Keras, 2021). Training details for the two experimental datasets are in Table 1. Training, test, and validation dataset spits were conducted randomly for the coconut trees dataset and spilt given in the original dataset were used for the Massachusetts buildings dataset.

Table 1: Training details for the two experimental datasets

|  | Massachusetts buildings dataset | Coconut trees dataset from the kingdom of Tonga (the south pacific) |
|---|---|---|
| Number of epochs | 100 | 50 |
| Batch size | 1 | 1 |
| Total training time | 101 minutes | 65 minutes |
| Input image size | 1280×1280 RGB | 1280×1280 RGB |
| Output image size | 1280×1280 images with Gaussian Kernels with 10 pixels of standard deviation | 1280×1280 images with Gaussian Kernels with 50 pixels of standard deviation |
| Total images | Total images: 151 (Training set: 137 images, Test set: 10 images and Validation set: 4 images) | Total images: 247 (Training set: 159 images, Test set: 58 images and Validation set: 30 images) |

### 3.4 Evaluation Criteria

Three metrics were used to evaluate the performance of the model: precision, recall, and F1-score (Equation 3, 4 and 5). First, True Positive (TP), True Negative (TN), False Positive (FP), and False Negative (FN) are calculated based on ML model performance. A TP and FP is an outcome where the model correctly and incorrectly predicts the positive class. Similarly, TN and FN is an outcome where the model correctly and incorrectly predicts the negative class. And then, we have calculated the following three metrics which are widely used to evaluate object localization tasks.

$$Precision = \frac{TP}{TP + FP} \tag{3}$$

$$Recall = \frac{TP}{TP + FN} \tag{4}$$

$$F1\ score = 2 \times \frac{Precision \times Recall}{Precision + Recall} \tag{5}$$

## 4. RESULTS

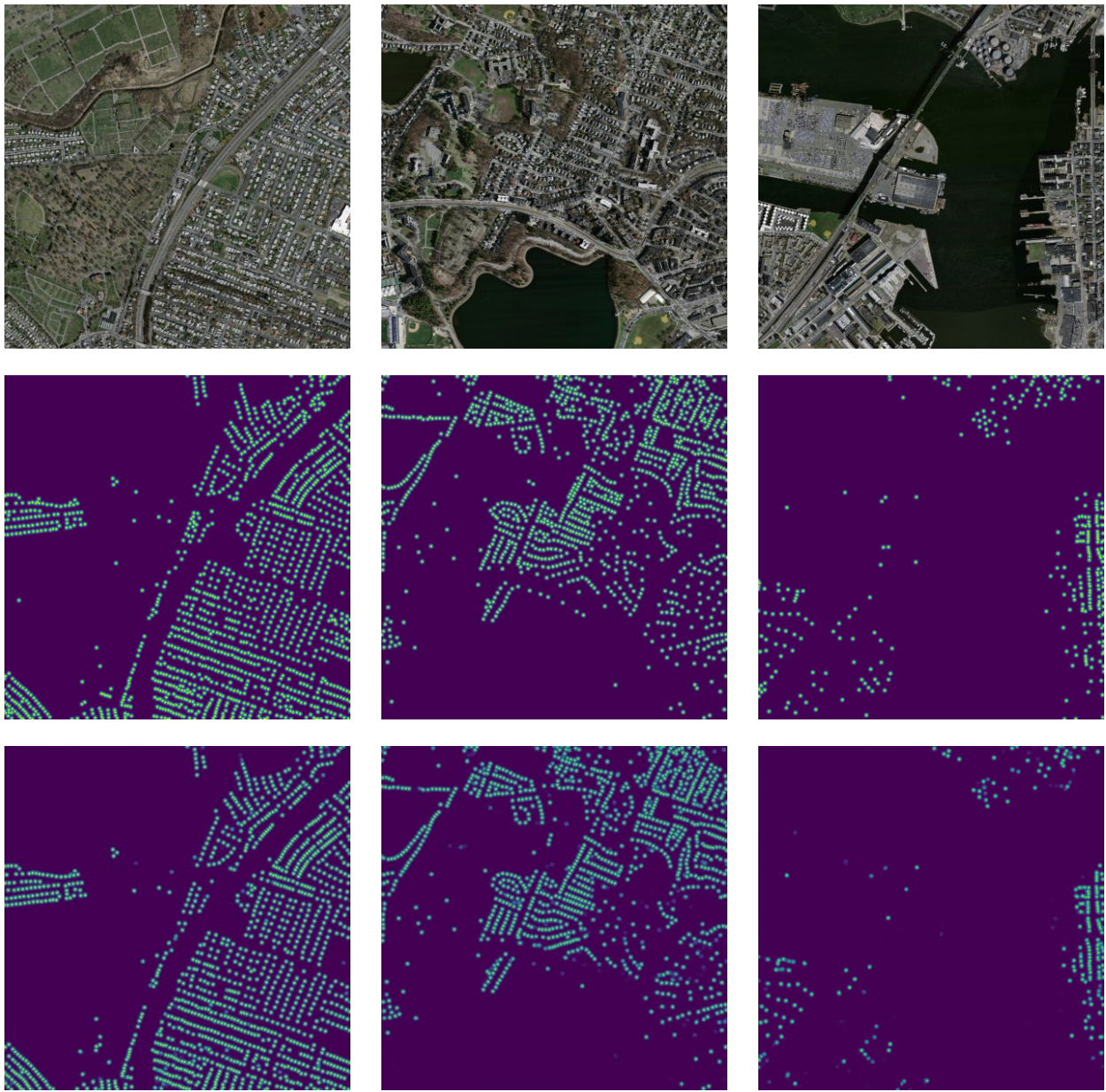

Figure 2: Sample experimental results for the test set of the Massachusetts buildings dataset. Input images are in the first row, ground truth images are in the second row, and predictions from our model are in the third row.

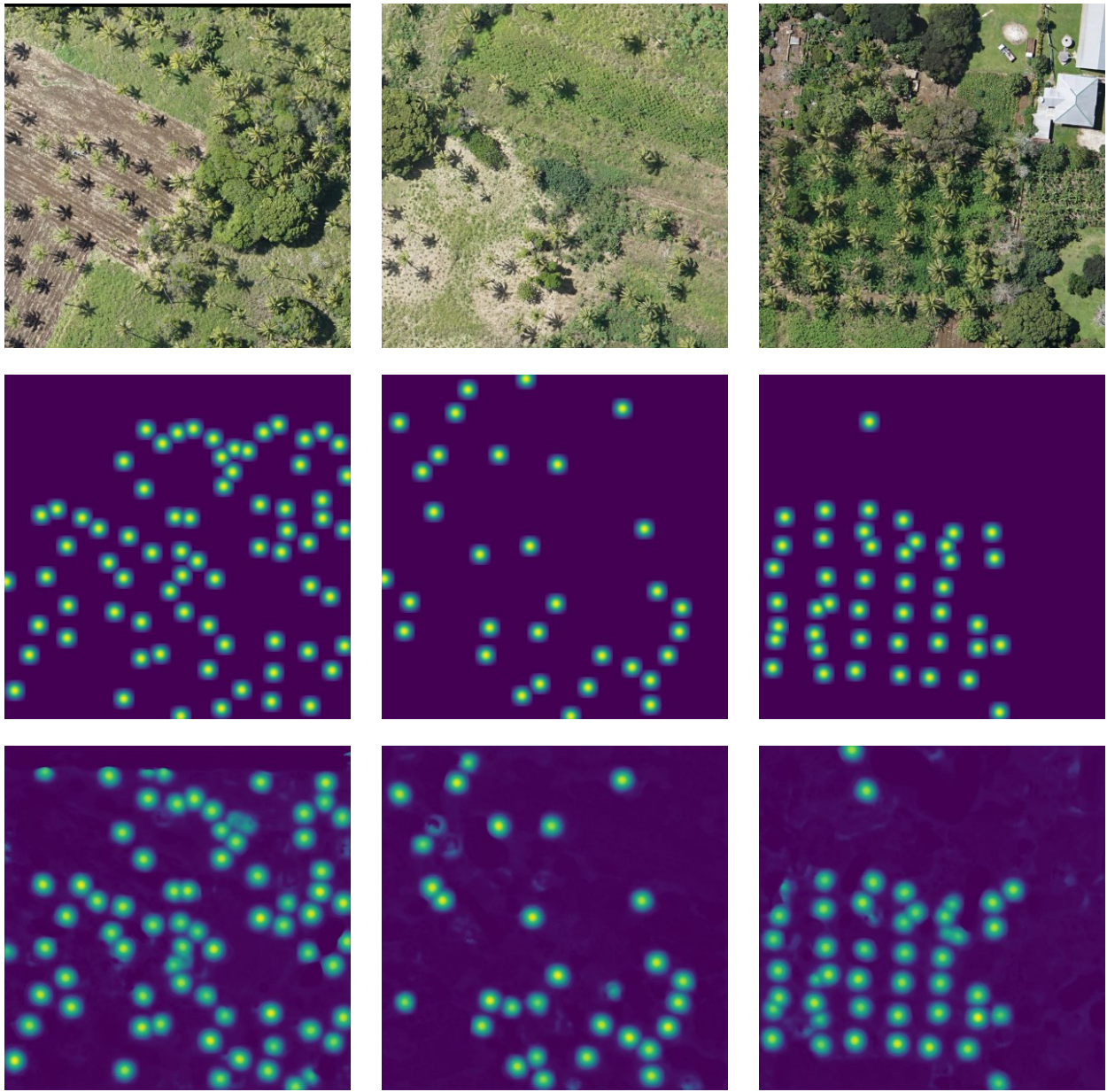

Figure 3: Sample experimental results for the test set of the coconut trees dataset from the kingdom of Tonga (the south pacific). Input images are in the first row, ground truth images are in the second row, and predictions from our model are in the third row.

Table 2: Accuracy assessment of two experiments over the test dataset.

| Dataset | Precision | Recall | F1 score |
|---|---|---|---|
| **Massachusetts buildings dataset** | 0.935 | 0.637 | 0.758 |
| **Coconut trees dataset from the kingdom of Tonga (the south pacific)** | 0.845 | 0.833 | 0.839 |

Figures 2 and 3 show sample results on the test dataset for the "Massachusetts buildings dataset" and the "Coconut trees dataset from the kingdom of Tonga (the south pacific)", respectively. Table 2 presents the results for the assessment of the accuracy for two experiments under three different evaluation criteria (precision, recall, and F1 score).

High F1 scores of 0.758 and 0.839 were achieved for building and coconut tree centroid mapping experiments, respectively. In both experiments, high precision values (0.935 and 0.845) were obtained, while recall values were relatively low (0.637 and 0.833), with the standard threshold of 0.5 on outputs of the network. Simply, this means, most of predictions from the model are correct with respect to the ground-truth data. These evaluation results are a considerable high, given the simplicity of the proposed approach.

## 5. CONCLUSION

In this research, we have evaluated the feasibility of using semantic segmentation networks such as U-Net for the object centroid mapping tasks without mapping bounding boxes or shapes of objects. This research proves that, by just adding Gaussian kernels around centroid points, we can use any semantic segmentation networks for the centroid mapping tasks in the context of satellite / aerial image analysis. This approach can be adapted to applications where only the locations of objects are important, allowing getting rid of unnecessary annotation work-loads for shapes and bounding boxes. And we have successfully evaluated our approach with two different datasets, achieving 0.758 and 0.839 F1 scores, with 0.935 and 0.845 precision values. As a future work, assessing the feasibility of using this approach for multi-class centroid estimation is highly important.